# A COMPARATIVE ANALYSIS OF A NEURAL NETWORK WITH CALCULATED WEIGHTS AND A NEURAL NETWORK WITH RANDOM GENERATION OF WEIGHTS BASED ON THE TRAINING DATASET SIZE

P.Sh. Geidarov

Institute of Control Systems, Azerbaijan National Academy of Sciences

**Abstract.** The paper discusses the capabilities of multilayer perceptron neural networks implementing metric recognition methods, for which the values of the weights are calculated analytically by formulas. Comparative experiments in training a neural network with pre-calculated weights and with random initialization of weights on different sizes of the MNIST training dataset are carried out. The results of the experiments show that a multilayer perceptron with pre-calculated weights can be trained much faster and is much more robust to the reduction of the training dataset.

**Introduction.** The implementation of artificial neural networks is known to require a very large training dataset, without which the training of a neural network becomes impossible. This complicates the implementation of neural networks for many applications. Even with such a training dataset available, the learning process itself also becomes lengthy and time-consuming. However, studies based on teaching small children show that the learning process in biological neural networks does not require such vast training datasets. Children learn with a small set of objects. Consequently, we see the relevance of creating new architectures of artificial neural networks, which could allow the neural network to learn with a smaller training dataset, similar to the way the biological neural networks do.

In this regard, it is worth recalling the architectures of neural networks based on metric recognition methods, which are proposed in [1, 2, 3]. These neural network architectures implement the algorithms of metric recognition methods [4] (*the nearest neighbor method, the k-nearest neighbor method, the method of potentials* and others) and are feedforward networks. Furthermore, the architectures of these networks also belong to the class of multilayer perceptrons (MLP). As shown on a MNIST dataset in the works [1, 2, 3, 5], the weights of these neural networks are calculated analytically by formulas using a small number of selected samples. The capability to calculate analytically the values of weights allows obtaining a functional neural network immediately, without the use of training algorithms and a training dataset. This capability is very similar to the biological brain's ability to recognize objects using a small training dataset. It was also shown in [6, 7] that the very process of calculating the neural network weights is performed very quickly, in fractions of seconds and minutes. It is shown in [7] that the resulting neural network can also be additionally trained by the *backpropagation* algorithm. It is also shown in [7] that the retraining process is much faster than in the classical scheme of neural network training by means of an initial random generation of weights.

The aim of this paper is to show the stability of calculatable neural networks in learning with a smaller training dataset, and thus to show that neural networks with calculated weights are more robust to the reduction of the size of the training dataset. In order to achieve this aim, we conduct experiments in training a neural network with calculated weights and with randomly generated weights on MNIST datasets, with different sizes of the training datasets, and perform a comparative analysis on the basis of the results of these experiments.

### 1. The main principles of a neural network based on metric recognition methods.

Recall the main principles of neural networks based on metric recognition methods:
The network structure (number of neurons, layers and connections) for these neural networks is determined strictly according to the diagram in Fig. 1, where the number of the first-layer neurons is $N(N-1)$, the number of the second-layer neurons is determined by the number of samples $N$, and the number of the third-layer neurons is determined by the number of patterns (classes) $N_{pat}$ of the problem. Each first-layer neuron performs a pairwise comparison of images of two samples (Fig. 2). To this end, a weight table is analytically calculated for each first-layer neuron. And each value of the weights of first-layer neurons is determined analytically, based on metric expressions of proximity, for instance, by the expression of the difference of squared Euclidean distances:

$$w_{c,r}^{(1)} = d_1^2 - d_2^2 = \left((c_1 - c_p)^2 + (r_1 - r_p)^2\right) - \left((c_2 - c_p)^2 + (r_2 - r_p)^2\right), \tag{1}$$

where ($c_1$, $r_1$) and ($c_2$, $r_2$) are the coordinates of the points (or cells of the weight table, Fig. 2b) to the nearest point (or cell) of the image of the sample with the coordinates ($c_p$, $r_p$) (Fig. 2a).

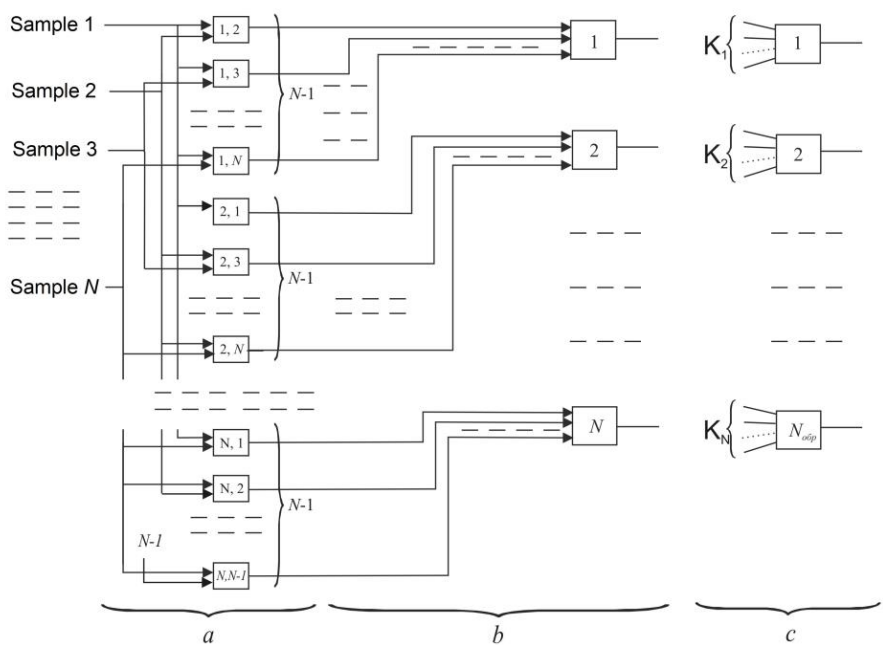

Fig. 1. The architecture of a neural network based on the nearest neighbor method for $N$ samples: (*a*) first layer, (*b*) second layer, (*c*) third layer of the neural network

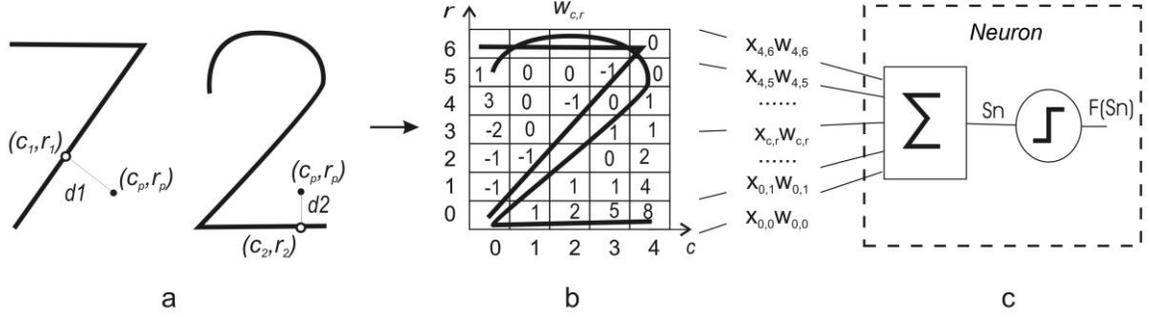

Fig. 2. (*a*) Distances $d_1$ and $d_2$ for the point ($c_p$, $r_p$), (*b*) A weight table for samples "2" and "7", (*c*) a neuron with a threshold activation function

The state function $Sn_{i,j}^{(1)}$ and the activation function $f\left(Sn_{i,j}^{(1)}\right)$ for each first-layer neuron are determined from expressions (2), (3):

$$Sn_{i,j}^{(1)} = \sum_{r=0}^{R}\sum_{c=0}^{C} x_{c,r} w_{c,r}^{(1)}, \tag{2}$$

$$\begin{cases} f\left(Sn_{i,j}^{(1)}\right) = 1, \text{ if } Sn_{i,j}^{(1)} > 0 \\ f\left(Sn_{i,j}^{(1)}\right) = 0, \text{ if } Sn_{i,j}^{(1)} < 0 \end{cases}, \tag{3}$$

where *C*, *R* are the numbers of the last column and the last row in the weight table (in Fig. 2*b*, *C*=4, *R*=6). According to (3), the threshold value of all first-layer neurons is 0.

In [3], a generalized algorithm is given for calculating the values of the weights of the second- and third-layer neurons, but in the simplest case the value of the weights for each input of the second- and third-layer neurons in Fig. 1 can be equal to 1.

$$w_{i,j}^{(2)} = w_k^{(3)} = 1 \tag{4}$$

The values of the state function $Sn_k^{(2)}$ and the activation function $f\left(Sn_k^{(2)}\right)$ for each *k*th neuron of the second layer are determined from expressions (5), (6):

$$Sn_k^{(2)} = \sum_{j=1, j \neq k}^{N} f\left(Sn_{k,j}^{(1)}\right), \tag{5}$$

$$\begin{cases} f\left(Sn_k^{(2)}\right) = 1, \text{ if } Sn_k^{(2)} \geq (N-1) = B^{(2)} \\ f\left(Sn_k^{(2)}\right) = 0, \text{ if } Sn_k^{(2)} < (N-1) = B^{(2)} \end{cases}. \tag{6}$$

Here, $B^{(2)} = N-1$ is the threshold value of a second-layer neuron, where *N* is the number of samples. If the output of the *k*th neuron of the second layer is $y_k^{(2)} = 1$, it means that the *k*th sample is the closest to the object $\overline{X}$ being recognized at the input of the neural network.

The third-layer neurons combine the outputs of the samples of one *k*th pattern into a single output $y_k^{(3)}$. The state function of the *k*th neuron of the third layer, $Sn_k^{(3)}$, and the activation function, $f\left(Sn_k^{(3)}\right)$, of a third-layer neuron are determined from the formulas:

$$Sn_k^{(3)} = \sum_{i \in k}^{K} f\left(Sn_i^{(2)}\right), \tag{7}$$

$$\begin{cases} f\left(Sn_k^{(3)}\right) = 1, \text{ if } Sn_k^{(3)} > 0 \\ f\left(Sn_k^{(3)}\right) = 0, \text{ if } Sn_k^{(3)} <= 0 \end{cases}. \tag{8}$$

The diagram in Fig. 1 implements a metric method—the nearest neighbor method. Fig. 3 shows the diagram of a neural network implementing another metric method—*the k-nearest neighbors method*.

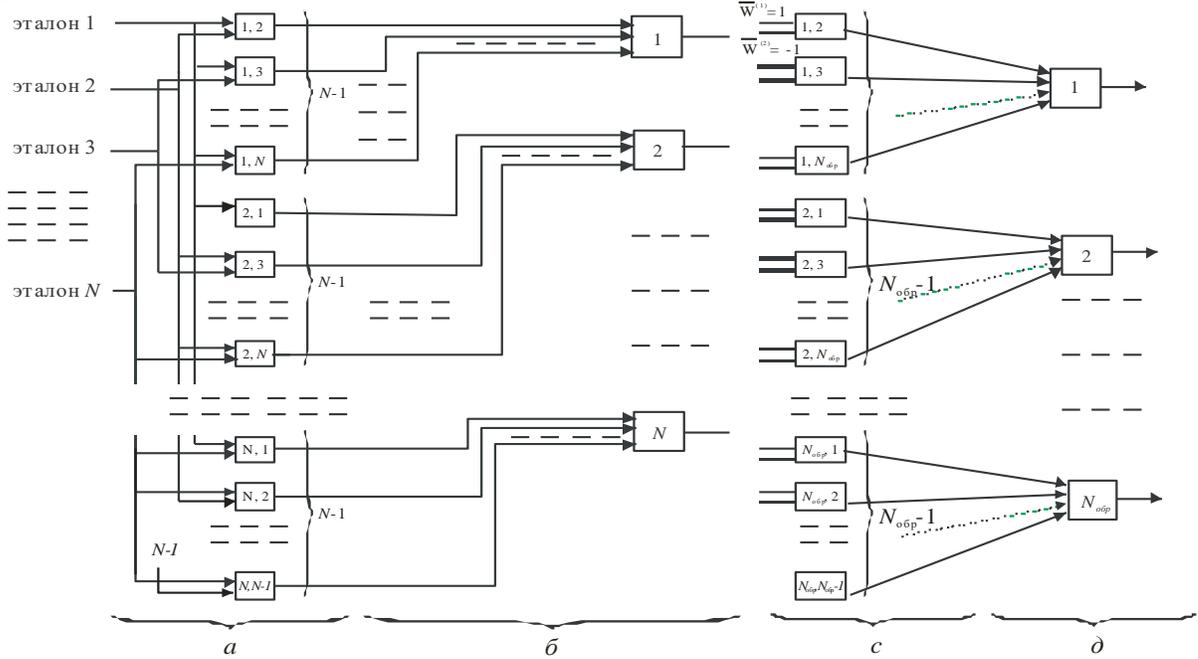

Fig. 3. A four-layer neural network implementing a metric recognition method— *the k-nearest neighbors method.*

The algorithm of the *nearest neighbor method* differs from that of the *k-nearest neighbors method* in that in the latter the proximity of the object being recognized is determined not to one, but to *n* samples, after which the final pattern to which more active samples out of *n* belong is determined. The object being recognized is assigned to this pattern. To implement this algorithm, the threshold value of the second-layer neurons in the diagram of the neural network in Fig. 1 is reduced from the value *N*-1 by the value $B^{(2)} = N - S$, which allows activating *S*=*n* outputs at once on the outputs of the second layer. In addition, the third layer in the diagram in Fig. 1 is changed and a fourth layer is added, Fig. 3. In the third-layer neuron we perform pairwise subtraction of the number of activated outputs of the second-layer neurons, whose outputs (or samples) belong to the two patterns being compared. The neuron's state function for the third pattern is determined from the expression:

$$Sn^{(3)}_{k,k1} = \sum_{i=0}^{N_k}\left(y_i^{(2)(k)} w_i^{(3)(k)}\right) - \sum_{i=0}^{N_{k1}}\left(y_i^{(2)(k1)} w_i^{(3)(k1)}\right), \qquad (9)$$

Here, *k, k1* are the names of two patterns for which the comparison of the number of active outputs in the neuron *(k, k1)* of the third layer is performed (Fig. 3). $N_k$, $N_{k1}$ are the numbers of samples for patterns *k* and *k1*. $y_i^{(2)(k)}$, $y_i^{(2)(k1)}$ are the values of the outputs of the second-layer neurons corresponding to patterns *k* and *k1*. The activation function of the third-layer neuron is determined from the expression:

$$\begin{cases} f\left(Sn^{(3)}_{k,k1}\right) = 1, \; if \; Sn^{(3)}_{k,k1} > 0 \\ f\left(Sn^{(3)}_{k,k1}\right) = 0, \; if \; Sn^{(3)}_{k,k1} \le 0 \end{cases}, \qquad (10)$$

Here, according to expression (9), the value of the weights of the third layer for the outputs of the second-layer neurons activating the outputs of the neurons of the *k*th pattern is equal to $w_k^{(3)(k)} = 1$, and for the outputs of the second-layer neurons activating the outputs of the neurons of the *k1*th pattern, it will be $w_k^{(3)(k1)} = -1$. The threshold value of all third-layer neurons is $w_k^{(3)(k1)} = -1$. The number of neurons in the third layer is $n^{(3)} = N_{pat}(N_{pat} - 1)$, where $N_{pat}$ is the number of recognized patterns (classes) of the problem.

The $k$th fourth-layer neuron sums the outputs of the third-layer neurons, in which the $k$th pattern is compared with other patterns. Here neuron output (12) is activated, in case if the value of the sum is $Sn_k^{(4)} \geq B^{(4)} = N_{pat} - 1$ (the threshold value of the fourth layer). The state function of the fourth-layer neuron is determined from expression (11):

$$Sn_k^{(4)} = \sum_{j=1, j \neq k}^{N_{pat}-1} f\left(Sn_{k,j}^{(3)}\right),  \quad (11)$$

$$\begin{cases} f\left(Sn_k^{(4)}\right) = 1, \ if \ Sn_k^{(4)} \geq (N_{pat} - 1) = B^{(4)} \\ f\left(Sn_k^{(4)}\right) = 0, \ if \ Sn_k^{(4)} < (N_{pat} - 1) = B^{(4)} \end{cases}. \quad (12)$$

The value of the weights for all input connections of the fourth-layer neurons is $w^{(4)} = 1$. The number of the fourth-layer neurons is equal to the number of patterns being recognized in the problem.

**2. Initial parameters and conditions of the experiment.**

In [7], two experiments in comparative analysis of the results of training a three-layer neural network with pre-calculated weights and with random generation of weights were conducted. For this purpose, we created a neural network based on metric recognition methods according to the diagram in Fig. 1 in a software module in the C++ Builder environment (Fig. 4), which allows creating and testing a neural network based on metric recognition methods according to the selected set of samples. In this study, the initial parameters and conditions for the experiments in training the neural network are kept the same as in [7].

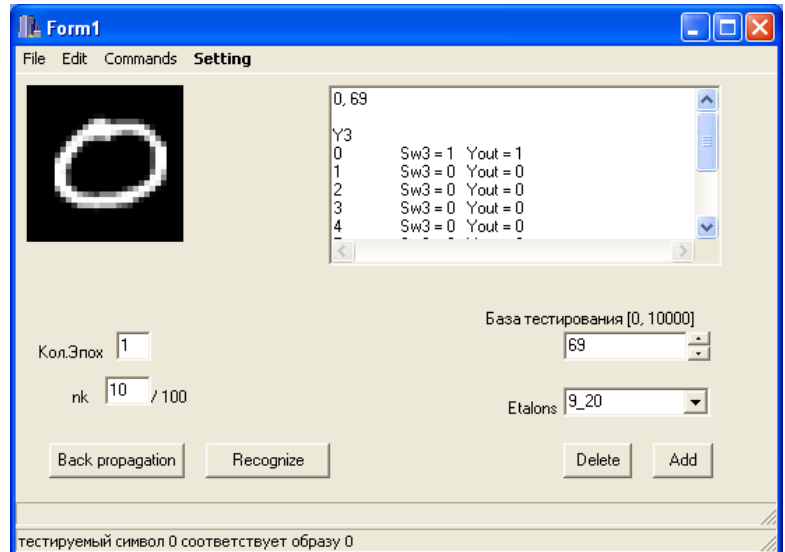

Fig. 4. The software module for recognizing and training images of the MNIST base, implemented in the C++ Builder environment

An arbitrarily selected set of 30 digits from the MNIST test dataset is used as standards (Fig. 5).

Fig. 5. Selected samples from the MNIST test set; the image name includes the name of the pattern and the number of the image in the MNIST dataset

For input elements we use a 28x56 binary input matrix, where the first part of the table (28x28) determines the pixels of the input image with values >150 as active and the pixels with values <150 as inactive. The second part of the binary table is mirror-opposite: pixels with values <150 are considered active, pixels with values >150 are considered inactive. The dimension of the weight tables for a first-layer neuron is also determined by the 28x56 dimension of the binary table. Formula (1) is used as an expression of the proximity measure, and the values of all weights are calculated by this formula. An example of a table of calculated weights for one first-layer neuron comparing, e.g., two samples, 2_172 and 5_102, from Fig. 5 is given in Fig. 6.

Fig. 6. Table of weights of the first layer for the neuron comparing Samples 2_172 and 5_102: (*a*) for the part of the binary matrix in which the light pixels of the image (>150) correspond to 1; (*b*) for the part of the binary matrix in which the darkened pixels of the image (<150) correspond to 1.

The number of the first-layer neurons is determined from the formula $n_1 = N(N - 1) = 30*29 = 870$ neurons; the number of the second-layer neurons equals the number of selected samples $n_2 = N = 30$; the number of the third-layer neurons equals the number of the patterns (digits) being recognized, $n3 = N_{pat} = 10$ neurons, (Fig. 1).

Fig. 7*ab* also shows the values of the weights and thresholds for all neuron connections for the second and third layers. Here the value of the weights of the connections of the second and third layers shown in Fig. 1 is 1, (4). Values 0 correspond to the values of the weights for the connections added to the neural network in Fig. 1, which makes the neural network in Fig. 1 fully-connected. The added connections here do not change the logic of the neural network operation.

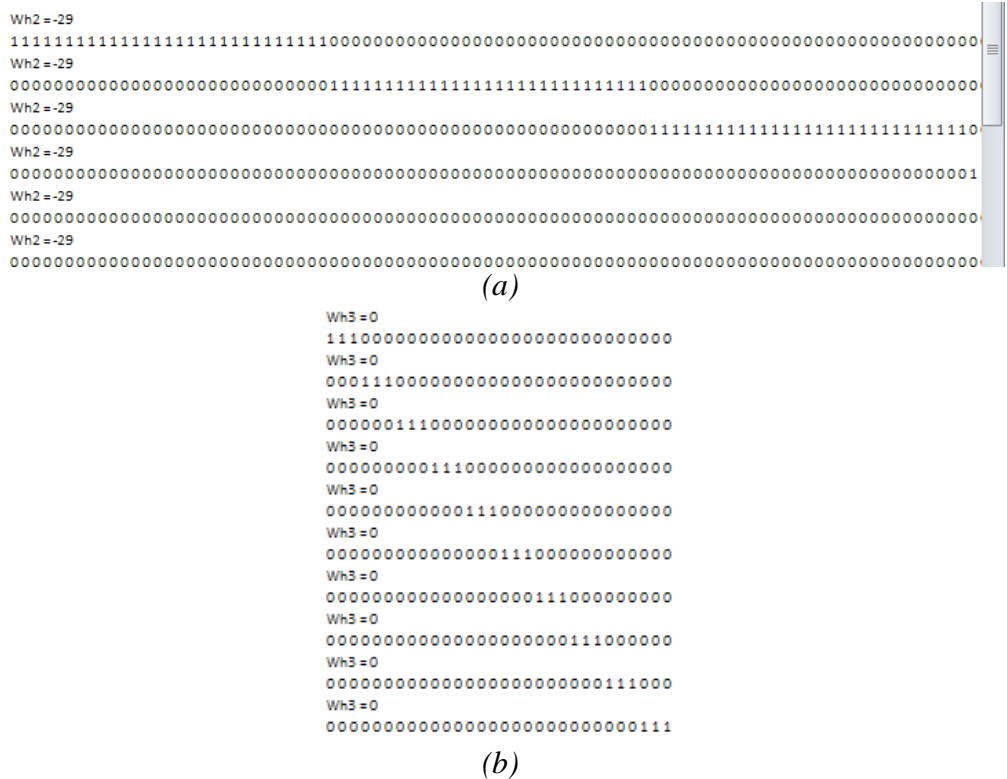

(a)

(b)

Fig. 7. (*a*) A fragment of weights for the second layer; (*b*) all weights of the third layer.

We used the sigmoid activation function as the activation function:
$$f(Sw) = \frac{1}{1+e^{-Sw}} \quad (13)$$

It should be said here that the threshold values of *B* given in (6, 8, 10, 12) are represented in (13) as values of the weights $w_0 = -B$. For instance, for this experiment the value of the second-layer neuron is $w_0^{(2)} = -B^{(2)} = -(N-1) = -(30-1) = -29$, (6).

Table 1 shows the overall results and the individual results for each recognized digit on the MNIST test dataset (10,000 images) after the analytical calculation of the weights. The final result without training the neural network was already 63%.

Note also that all the calculations described in this paper and in [7] were performed on the same computer, in the same software module shown in Fig. 4, and using the same three-layer neural network. The parameters of the *backpropagation* learning algorithm were also the same for all the experiments. Three epochs were used to train the neural network, the first two of which were trained at the rate $nk = 0.1$, and the last one was trained at the rate $nk = 0.02$. The stochastic

*backpropagation* learning algorithm was used. The learning error $S_{err}$ was calculated for each epoch using the formula:

$$S_{err} = \frac{1}{2} \sum_{i=0}^{P} \sum_{k=0}^{N_{pat}} \left( y_k^{(corr)} - f(Sn_k^{(3)}) \right), \qquad (14)$$

where $y_k^{(corr)}$ is the adopted correct value of the *k*th output of the third layer, for the active output $y_k^{(corr)} = 0.7$, and for the inactive $y_k^{(corr)} = 0{,}2$; *P* is the number of incorrectly identified images from the MNIST training set for which the weights were corrected by the backpropagation algorithm.

Table 1. Recognition results for the MNIST test set (10,000 images) without training. Here, *sj* is the number of correctly recognized digits of pattern *j* in the MNIST test set, *ij* is the total number of digits of pattern *j* in the MNIST test set, and *pj* is the percentage of correctly recognized digits for pattern *j*.

| | | |
|---|---|---|
| s0 = 834 | i0 = 980 | p0 = 85% |
| s1 = 968 | i1 = 1135 | p1 = 85% |
| s2 = 530 | i2 = 1032 | p2 = 51% |
| s3 = 454 | i3 = 1010 | p3 = 44% |
| s4 = 410 | i4 = 982 | p4 = 41% |
| s5 = 411 | i5 = 892 | p5 = 46% |
| s6 = 586 | i6 = 958 | p6 = 61% |
| s7 = 556 | i7 = 1028 | p7 = 54% |
| s8 = 773 | i8 = 974 | p8 = 79% |
| s9 = 750 | i9 = 1009 | p9 = 74% |
| **Total** | | |
| **s = 6272** | **i = 10000** | **p = 63%** |

Thus, at this point we can see that with only 30 samples, a multilayer perceptron can already recognize most (63%) of the MNIST test dataset, with just half a second ($t_{construct}$ = 0.5469 s) spent on the calculation of the weights.

Recall also that in the first experiment in [7] we trained the obtained neural network with already calculated weights. In the second experiment with the same neural network, the neural network was trained in the classical way, that is, without using the pre-calculation of the weights. For the second experiment, the values of the weights were determined arbitrarily. The comparison results are shown in Table 2 for the MNIST training dataset of 60,000 images, and in Table 3 for the MNIST test dataset of 10,000 images of digits. According to the results in Tables 2 and 3, for the case of training a neural network with pre-calculated weights, the number of correctly recognized digits in all training epochs is higher than for the training of a neural network with random generation of weights. Moreover, the process of retraining an analytically calculated neural network takes 34% less time, $p_{speedup}$ = (499-329)*100/499 ≈ 34%.

Table 2. Comparing the results of training the neural network on the MNIST training set (60,000 images) for each training epoch.

| Epoch no. | Learning rate | Training a neural network with pre-calculated weights | | | | Training a neural network with random initialization of the weights in the range [−0.5; 0.5] | | | |
|---|---|---|---|---|---|---|---|---|---|
| | | Number of recognized images | % of recognized images | $S_{err}$ | time, min | Number of recognized images | % of recognized images | $S_{err}$ | time, min |
| 1 | 0.1 | 43932 | 73% | 1199 | 159 | 35370 | 59% | 1935 | 256 |
| 2 | 0.1 | 49748 | 83% | 737 | 98 | 46033 | 76% | 1051 | 139 |
| 3 | 0.02 | 52285 | 87% | 545 | 72 | 49195 | 82% | 784 | 104 |
| **Total training time in minutes** | | | | | **329** | **Total training time in minutes** | | | **499** |

Table 3. Comparing the results of training the neural network on the MNIST test set (10,000 images) for each training epoch, where the neural network has been trained on a training set of 60,000 images.

| Epoch no. | Learning rate | With pre-calculated weights | | With initial random initialization of weights | |
|---|---|---|---|---|---|
| | | number of recognized images | % | number of recognized images | % |
| 1 | 0.1 | 9145 | 91.45% | 8894 | 88.94% |
| 2 | 0.1 | 9282 | 92.82% | 9116 | 91.16% |
| **3** | **0.02** | **9449** | **94.49%** | **9256** | **92.56%** |

**3. Comparative experiments with a smaller training set.**

To test the stability of a neural network with calculated weight values and with random generation of weights, we reduce the MNIST training set and conduct the same comparative experiments with the same initial conditions. Table 4 shows the results of training a neural network with calculated and random initial weights trained on the MNIST database with a training set of 40,000 images. From the results in Table 4 we can see that in all three training epochs the number of recognized images is higher for the neural network with calculated weight values, and the total training time of the neural network with calculated weights is 38% less, ($p_{speedup}$ = (471-293)*100/471 ≈ 38%). In Tables 5, 6, 7 the trained neural network is tested on the MNIST test dataset (10,000 images). The results of Tables 5, 6, 7 also show that the effectiveness of testing on a test dataset after each training epoch is also higher for a neural network with calculated weight values.

Table 8 also shows the results of training a neural network with calculated and random initial weights trained on the MNIST database with a training set of 20,000 images. The results in Tables 9, 10, 11 also make it clear that in all three training epochs the number of recognized images is higher for the neural network with calculated weight values, and the total training time of the neural network with calculated weight values is 34% less, ($p_{speedup}$ = (298-196)*100/298 ≈ 34%). Tables 9, 10, 11 also show the results of tests of the neural network already trained on the MNIST test set (10,000 images). It is clear from the results in the tables, that the number of correctly recognized symbols in the test set after each training epoch is also higher for the neural network with the calculated weights.

Table 4. Results of training a neural network with calculated weights and with random generation of weights on a dataset of 40,000 images.

| Epoch no. | Learning rate | Training a neural network with pre-calculated weights | | | | Training a neural network with random initialization of the weights in the range [−0.5; 0.5] | | | |
|---|---|---|---|---|---|---|---|---|---|
| | | Number of recognized images | % of recognized images | $S_{err}$ | time, min | Number of recognized images | % of recognized images | $S_{err}$ | time, min |
| 1 | 0.1 | 32815 | 82% | 765.7 | 137.6 | 27346 | 68.3% | 1314.5 | 228.6 |
| 2 | 0.1 | 35705 | 89.2% | 411 | 88.47 | 32578 | 81.44% | 729.25 | 140.4 |
| 3 | 0.02 | 36899 | 92.2% | 295.4 | 67.5 | 34851 | 87.1% | 498.7 | 102.8 |
| | | **Total training time in minutes** | | | **293,57** | **Total training time in minutes** | | | **471,8** |

Table 5. Results of testing on a test set of 10,000 images for a neural network with calculated weights retrained on a training dataset of 40,000 images

| Epoch 1 | | | Epoch 2 | | | Epoch 3 | | |
|---|---|---|---|---|---|---|---|---|
| s0 = 898 | i0 = 980 | p0 = 91% | s0 = 937 | i0 = 980 | p0 = 95% | s0 = 956 | i0 = 980 | p0 = 97% |
| s1 = 1093 | i1 = 1135 | p1 = 96% | s1 = 1108 | i1 = 1135 | p1 = 97% | s1 = 1106 | i1 = 1135 | p1 = 97% |
| s2 = 827 | i2 = 1032 | p2 = 80% | s2 = 839 | i2 = 1032 | p2 = 81% | s2 = 935 | i2 = 1032 | p2 = 90% |
| s3 = 815 | i3 = 1010 | p3 = 80% | s3 = 815 | i3 = 1010 | p3 = 80% | s3 = 944 | i3 = 1010 | p3 = 93% |
| s4 = 825 | i4 = 982 | p4 = 84% | s4 = 920 | i4 = 982 | p4 = 93% | s4 = 931 | i4 = 982 | p4 = 94% |
| s5 = 821 | i5 = 892 | p5 = 92% | s5 = 854 | i5 = 892 | p5 = 95% | s5 = 786 | i5 = 892 | p5 = 88% |
| s6 = 865 | i6 = 958 | p6 = 90% | s6 = 831 | i6 = 958 | p6 = 86% | s6 = 895 | i6 = 958 | p6 = 93% |
| s7 = 898 | i7 = 1028 | p7 = 87% | s7 = 922 | i7 = 1028 | p7 = 89% | s7 = 912 | i7 = 1028 | p7 = 88% |
| s8 = 842 | i8 = 974 | p8 = 86% | s8 = 838 | i8 = 974 | p8 = 86% | s8 = 862 | i8 = 974 | p8 = 88% |
| s9 = 847 | i9 = 1009 | p9 = 83% | s9 = 846 | i9 = 1009 | p9 = 83% | s9 = 867 | i9 = 1009 | p9 = 85% |
| **Total** | | | **Total** | | | **Total** | | |
| **8731** | **10000** | **87%** | **8910** | **10000** | **89%** | **9194** | **10000** | **92%** |

Table 6. Results of testing on a test set of 10,000 images for a neural network with random generation of weights trained on a training dataset of 40,000 images

| Epoch 1 | | | Epoch 2 | | | Epoch 3 | | |
|---|---|---|---|---|---|---|---|---|
| s0 = 402 | i0 = 980 | p0 = 41% | s0 = 370 | i0 = 980 | p0 = 37% | s0 = 867 | i0 = 980 | p0 = 88% |
| s1 = 995 | i1 = 1135 | p1 = 87% | s1 = 969 | i1 = 1135 | p1 = 85% | s1 = 1078 | i1 = 1135 | p1 = 94% |
| s2 = 287 | i2 = 1032 | p2 = 27% | s2 = 263 | i2 = 1032 | p2 = 25% | s2 = 788 | i2 = 1032 | p2 = 76% |
| s3 = 251 | i3 = 1010 | p3 = 24% | s3 = 443 | i3 = 1010 | p3 = 43% | s3 = 739 | i3 = 1010 | p3 = 73% |
| s4 = 772 | i4 = 982 | p4 = 78% | s4 = 908 | i4 = 982 | p4 = 92% | s4 = 856 | i4 = 982 | p4 = 87% |
| s5 = 833 | i5 = 892 | p5 = 93% | s5 = 844 | i5 = 892 | p5 = 94% | s5 = 772 | i5 = 892 | p5 = 86% |
| s6 = 728 | i6 = 958 | p6 = 75% | s6 = 733 | i6 = 958 | p6 = 76% | s6 = 795 | i6 = 958 | p6 = 82% |
| s7 = 805 | i7 = 1028 | p7 = 78% | s7 = 810 | i7 = 1028 | p7 = 78% | s7 = 939 | i7 = 1028 | p7 = 91% |
| s8 = 867 | i8 = 974 | p8 = 89% | s8 = 847 | i8 = 974 | p8 = 86% | s8 = 886 | i8 = 974 | p8 = 90% |
| s9 = 721 | i9 = 1009 | p9 = 71% | s9 = 682 | i9 = 1009 | p9 = 67% | s9 = 813 | i9 = 1009 | p9 = 80% |
| **Total** | | | **Total** | | | **Total** | | |
| **6661** | **10000** | **66%** | **6869** | **10000** | **68%** | **8533** | **10000** | **85%** |

Table 7. Comparing the results of training the neural network for 40,000 images, with a testing on the MNIST test set (10,000 images) for each training epoch.

| Epoch no. | Learning rate | With pre-calculated weights | | With initial random initialization of weights | |
|---|---|---|---|---|---|
| | | number of recognized images | % | number of recognized images | % |
| 1 | 0.1 | 8731 | 87.31% | 6661 | 66.61% |
| 2 | 0.1 | 8910 | 89.10% | 6869 | 68.69% |
| 3 | 0.02 | **9194** | **91.94%** | **8533** | **85.33%** |

Table 8. Results of training a neural network with calculated weights and with random generation of weights for a dataset of 20,000 images.

| Epoch no. | Learning rate | Training the neural network with pre-calculated weights | | | | Training the neural network with random initialization of the weights in the range $[-0.5; 0.5]$ | | | |
|---|---|---|---|---|---|---|---|---|---|
| | | Number of recognized images | % of recognized images | $S_{err}$ | time, min | Number of recognized images | % of recognized images | $S_{err}$ | time, min |
| 1 | 0.1 | 15564 | 77.82% | 498.87 | 90.27 | 12497 | 62.48% | 798.8 | 142.23 |
| 2 | 0.1 | 17412 | 87.06% | 254.61 | 58.8 | 15552 | 77.76% | 437.9 | 90.39 |
| 3 | 0.02 | 18098 | 90.49% | 183.77 | 47.13 | 17015 | 85.07% | 290.07 | 65.51 |
| **Total training time in minutes** | | | | | **196,2** | **Total training time in minutes** | | | **298,13** |

Table 9. Results of testing on a test set of 10,000 images for a neural network with calculated weights retrained on a training set of 20,000 images.

| Epoch 1 | | | Epoch 2 | | | Epoch 3 | | |
|---|---|---|---|---|---|---|---|---|
| $s_0 = 945$ | $i_0 = 980$ | $p_0 = 96\%$ | $s_0 = 929$ | $i_0 = 980$ | $p_0 = 94\%$ | $s_0 = 939$ | $i_0 = 980$ | $p_0 = 95\%$ |
| $s_1 = 1111$ | $i_1 = 1135$ | $p_1 = 97\%$ | $s_1 = 1116$ | $i_1 = 1135$ | $p_1 = 98\%$ | $s_1 = 1106$ | $i_1 = 1135$ | $p_1 = 97\%$ |
| $s_2 = 851$ | $i_2 = 1032$ | $p_2 = 82\%$ | $s_2 = 848$ | $i_2 = 1032$ | $p_2 = 82\%$ | $s_2 = 907$ | $i_2 = 1032$ | $p_2 = 87\%$ |
| $s_3 = 773$ | $i_3 = 1010$ | $p_3 = 76\%$ | $s_3 = 760$ | $i_3 = 1010$ | $p_3 = 75\%$ | $s_3 = 876$ | $i_3 = 1010$ | $p_3 = 86\%$ |
| $s_4 = 837$ | $i_4 = 982$ | $p_4 = 85\%$ | $s_4 = 907$ | $i_4 = 982$ | $p_4 = 92\%$ | $s_4 = 882$ | $i_4 = 982$ | $p_4 = 89\%$ |
| $s_5 = 663$ | $i_5 = 892$ | $p_5 = 74\%$ | $s_5 = 795$ | $i_5 = 892$ | $p_5 = 89\%$ | $s_5 = 794$ | $i_5 = 892$ | $p_5 = 89\%$ |
| $s_6 = 888$ | $i_6 = 958$ | $p_6 = 92\%$ | $s_6 = 886$ | $i_6 = 958$ | $p_6 = 92\%$ | $s_6 = 903$ | $i_6 = 958$ | $p_6 = 94\%$ |
| $s_7 = 859$ | $i_7 = 1028$ | $p_7 = 83\%$ | $s_7 = 916$ | $i_7 = 1028$ | $p_7 = 89\%$ | $s_7 = 897$ | $i_7 = 1028$ | $p_7 = 87\%$ |
| $s_8 = 850$ | $i_8 = 974$ | $p_8 = 87\%$ | $s_8 = 851$ | $i_8 = 974$ | $p_8 = 87\%$ | $s_8 = 859$ | $i_8 = 974$ | $p_8 = 88\%$ |
| $s_9 = 867$ | $i_9 = 1009$ | $p_9 = 85\%$ | $s_9 = 766$ | $i_9 = 1009$ | $p_9 = 75\%$ | $s_9 = 915$ | $i_9 = 1009$ | $p_9 = 90\%$ |
| **Total** | | | **Total** | | | **Total** | | |
| **8644** | **10000** | **86%** | **8774** | **10000** | **87%** | **9078** | **10000** | **90%** |

Table 10. Results of testing on a test set of 10,000 images for a neural network with random generation of the weights and trained on a training set of 20,000 images.

| Epoch 1 | | | Epoch 2 | | | Epoch 3 | | |
|---|---|---|---|---|---|---|---|---|
| s0 = 687 | i0 = 980 | p0 = 70% | s0 = 851 | i0 = 980 | p0 = 86% | s0 = 849 | i0 = 980 | p0 = 86% |
| s1 = 1051 | i1 = 1135 | p1 = 92% | s1 = 1092 | i1 = 1135 | p1 = 96% | s1 = 1067 | i1 = 1135 | p1 = 94% |
| s2 = 503 | i2 = 1032 | p2 = 48% | s2 = 678 | i2 = 1032 | p2 = 65% | s2 = 847 | i2 = 1032 | p2 = 82% |
| s3 = 429 | i3 = 1010 | p3 = 42% | s3 = 564 | i3 = 1010 | p3 = 55% | s3 = 663 | i3 = 1010 | p3 = 65% |
| s4 = 599 | i4 = 982 | p4 = 60% | s4 = 560 | i4 = 982 | p4 = 57% | s4 = 357 | i4 = 982 | p4 = 36% |
| s5 = 778 | i5 = 892 | p5 = 87% | s5 = 788 | i5 = 892 | p5 = 88% | s5 = 804 | i5 = 892 | p5 = 90% |
| s6 = 729 | i6 = 958 | p6 = 76% | s6 = 761 | i6 = 958 | p6 = 79% | s6 = 835 | i6 = 958 | p6 = 87% |
| s7 = 651 | i7 = 1028 | p7 = 63% | s7 = 691 | i7 = 1028 | p7 = 67% | s7 = 874 | i7 = 1028 | p7 = 85% |
| s8 = 744 | i8 = 974 | p8 = 76% | s8 = 800 | i8 = 974 | p8 = 82% | s8 = 825 | i8 = 974 | p8 = 84% |
| s9 = 880 | i9 = 1009 | p9 = 87% | s9 = 931 | i9 = 1009 | p9 = 92% | s9 = 904 | i9 = 1009 | p9 = 89% |
| **Total** | | | **Total** | | | **Total** | | |
| **7051** | **10000** | **70%** | **7716** | **10000** | **77%** | **8025** | **10000** | **80%** |

Table 11. Comparing the results of training the neural network on 20,000 images, with a testing on the MNIST test set (10,000 images) for each training epoch.

| Epoch no. | Learning rate | With pre-calculated weights | | With initial random initialization of weights | |
|---|---|---|---|---|---|
| | | number of recognized images | % | number of recognized images | % |
| 1 | 0.1 | 8644 | 86.44% | 7051 | 70.51% |
| 2 | 0.1 | 8774 | 87.74% | 7716 | 77.16% |
| 3 | 0.02 | **9078** | **90.78%** | **8025** | **80.25%** |

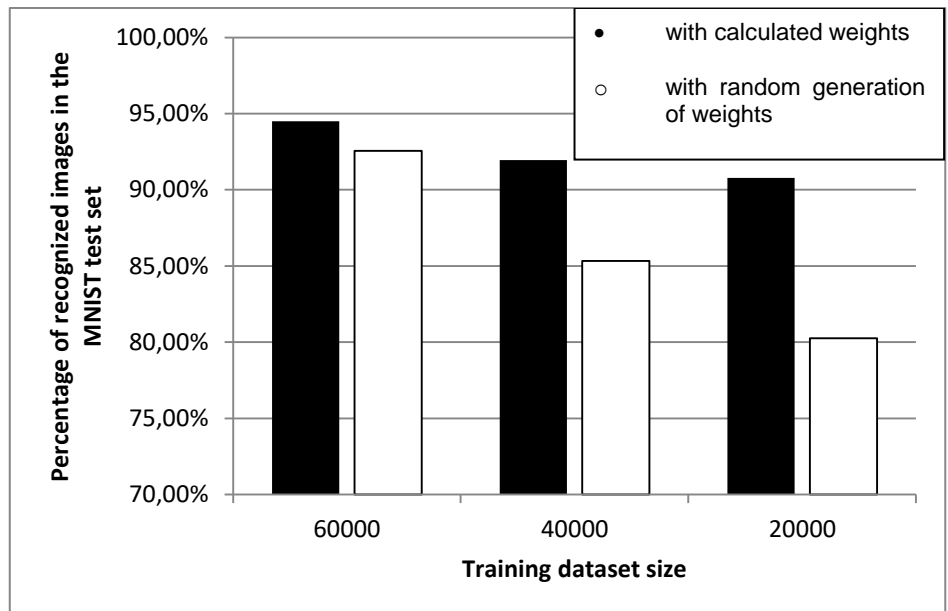

Fig. 8. The diagram of the percentage of recognized digits in the MNIST test set for a neural network with calculated weights and with random generation of weights vs the training set size.

The diagram in Fig. 8 shows the values of the number of correctly recognized digits from the MNIST test dataset (10,000 images) for a neural network with pre-calculated weights and with random initialization of weights relative to the size of the training set used. In the diagram in Fig. 8, we can also observe that the neural network with calculated weights is significantly more robust to the reduction of the training set size. Even with the training set of 20,000 characters, the percentage of recognized characters from the MNIST test dataset (10,000 images) is still more than 90%, while a neural network with random initialization of weights is extremely unstable to any decrease in the training set, and with 20,000 training characters the percentage of recognized characters drops to 80%.

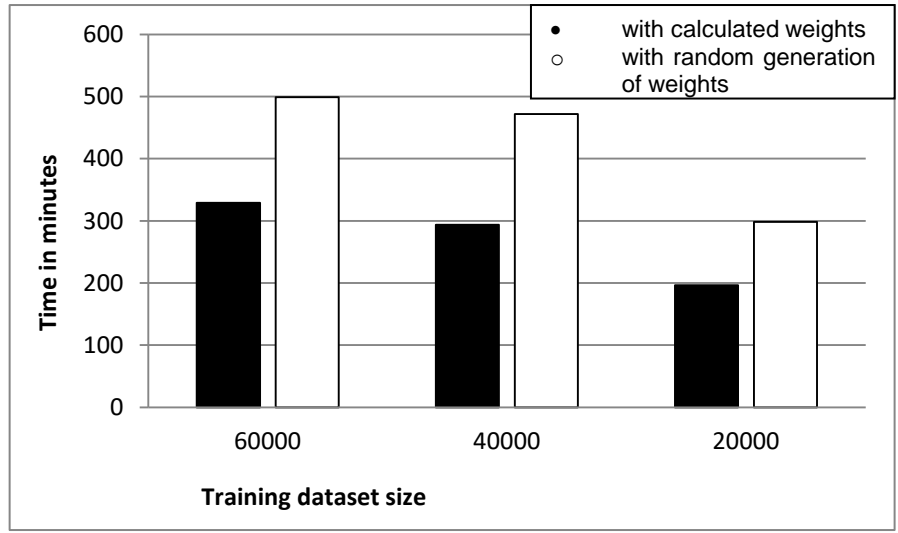

Fig. 9. The diagram of the time spent on training a neural network with calculated weights and with random generation of weights vs the training set size.

Fig. 9 shows a diagram of the time spent on training a neural network with calculated weights and with random generation of weights relative to the training set size. We can also see from the diagram in Fig. 9 that the training procedure is faster for the neural network with pre-calculated weights with different amounts of training data of the MNIST dataset. It can also be said that when using a neural network scheme that implements another, stronger metric method of recognition, such as the *k-nearest neighbors method* (Fig. 3), the percentage of correctly identified characters immediately after the calculation of the weights will be higher than 63% given in Table 1. Accordingly, the results given above in Tables 2-11 will also be better.

From Tables 3, 10 and the diagram in Fig. 8 we can see that after three training epochs of the neural network, the number of correctly identified characters for a multilayer perceptron with calculated weights with a training set size of 20,000 characters is approximately equal to the number of correctly identified characters for a neural network with random initial generation of weights with a training set size of 60,000 characters (9078 ≈ 9256). We can also see from Tables 8 and 2, that the time required for training a neural network with pre-calculated weights on a training set of 20,000 characters is 196 minutes, while the time required for training a neural network with initial random generation of weights on a training set of 60,000 characters is 499 minutes. Thus we can say that the procedure of training a multilayer perceptron with pre-calculated weights further speeds up by the value $p_{speedup} = (499-196)*100/499 \approx 60.72\%$.

**Conclusion.** Based on the above experiments, we arrive at the following conclusions:

1. The weights of a multilayer perceptron can be immediately calculated analytically using a small set of arbitrarily selected samples (Fig. 5). Thus, we get a functional neural network right away (Tab.1), which can also be additionally trained by classical training algorithms (Tables 2, 4, 8).

2. The above experiments and the results obtained in Tables 4-11 have once again confirmed that the procedure of retraining a multilayer perceptron with pre-calculated weights is much faster than the training of a neural network with initial random generation of weights, and the result of the comparative experiment given in [7] was not random but logical.

3. The experiments have shown that a neural network with analytically calculated weights is much more robust to the reduction of the training dataset of a multilayer perceptron. Thus, the process of retraining a neural network can use a smaller training set and thus further speed up the learning of a multilayer perceptron.

4. We can also say that even better results can be achieved by using better metric recognition methods (e.g., the *k-nearest neighbors algorithm*, Fig. 3) and a larger sample set in multilayer perceptron neural network schemes.